\documentclass[letterpaper, 10 pt, conference]{ieeeconf}  

\IEEEoverridecommandlockouts                              

\usepackage{graphicx} 
\usepackage{amsmath} 
\usepackage{amssymb}  
\usepackage{amsfonts}
\usepackage{tabularx}
\usepackage[table]{xcolor}

\title{\LARGE \bf
Clone Swarms: Learning to Predict and Control Multi-Robot Systems by Imitation
}

\author{Siyu Zhou$^{1}$, Mariano Phielipp$^{2}$, Jorge A. Sefair$^{3}$, Sara I. Walker$^{4}$ and Heni Ben Amor$^{3}$
\thanks{$^{1}$Department of Physics, Arizona State University
        {\tt\small siyu.zhou@asu.edu}
        }%
\thanks{$^{2}$Intel Corporation
        {\tt\small mariano.j.phielipp@intel.com}
        }%
\thanks{$^{3}$School of Computing, Informatics, and Decision Systems Engineering, Arizona State University
        {\tt\small \{jorge.sefair, hbenamor\}@asu.edu}
        }
\thanks{$^{4}$School of Earch and Space Exploration, Arizona State University
        {\tt\small sara.i.walker@asu.edu}
        }
}

\begin{document}

\maketitle

\thispagestyle{empty}
\pagestyle{empty}

\begin{abstract}
In this paper, we propose SwarmNet -- a neural network architecture that can learn to predict and imitate the behavior of an observed swarm of agents in a centralized manner. Tested on artificially generated swarm motion data, the network achieves high levels of prediction accuracy and imitation authenticity. We compare our model to previous approaches for modelling interaction systems and show how modifying components of other models gradually approaches the performance of ours. Finally, we also discuss an extension of SwarmNet that can deal with nondeterministic, noisy, and uncertain environments, as often found in robotics applications.
\end{abstract}

\section{Introduction}
\label{sec:intro}
Multi-Robot Systems (MRS)~\cite{arai2002advances} describe groups of robotic agents that collectively perform complex tasks in a distributed and parallel manner through repeated interactions among each other and the environment. Such systems have attracted considerable attention in recent years with remarkable successes in a number of application domains, including defense, agriculture, logistics, disaster management, and entertainment. In particular, today's fast-paced online economy is largely fuelled by tens of thousands of warehouse robots that transport millions of items across fulfillment centers all over the world.

Despite this progress, programming groups of robots to perform a joint task is still considered a complex, time-consuming, and extremely challenging endeavour. One prominent formalism for the specification of MRS is based on the identification of cost functions~\cite{lagoudakis2004simple} governing the group behavior. However, this approach is not intuitive and requires a deep understanding of complex theoretical concepts across a number of mathematical fields, e.g., graph theory, manifold theory, nonlinear optimization, etc. In addition, the real-world ramifications of even small changes in a given cost function are extremely difficult to foresee. An alternative approach is to provide a set of building-block behaviors~\cite{Reynolds1999} that are combined to produce overall group strategies, e.g., formation or area covering. An important aspect in this regard is the notion of \emph{emergence} in which simple rules interact with each other to generate a whole that is more complex than the sum of its parts. Higher levels of complexity emerge as a result of repeated simple local interactions. This transition in complexity due to emergent properties is difficult to predict and introduces substantial challenges in the design of the right set of building-block behaviors. 

\begin{figure}[t!] 
    \centering
    \includegraphics[width = 0.4\textwidth]{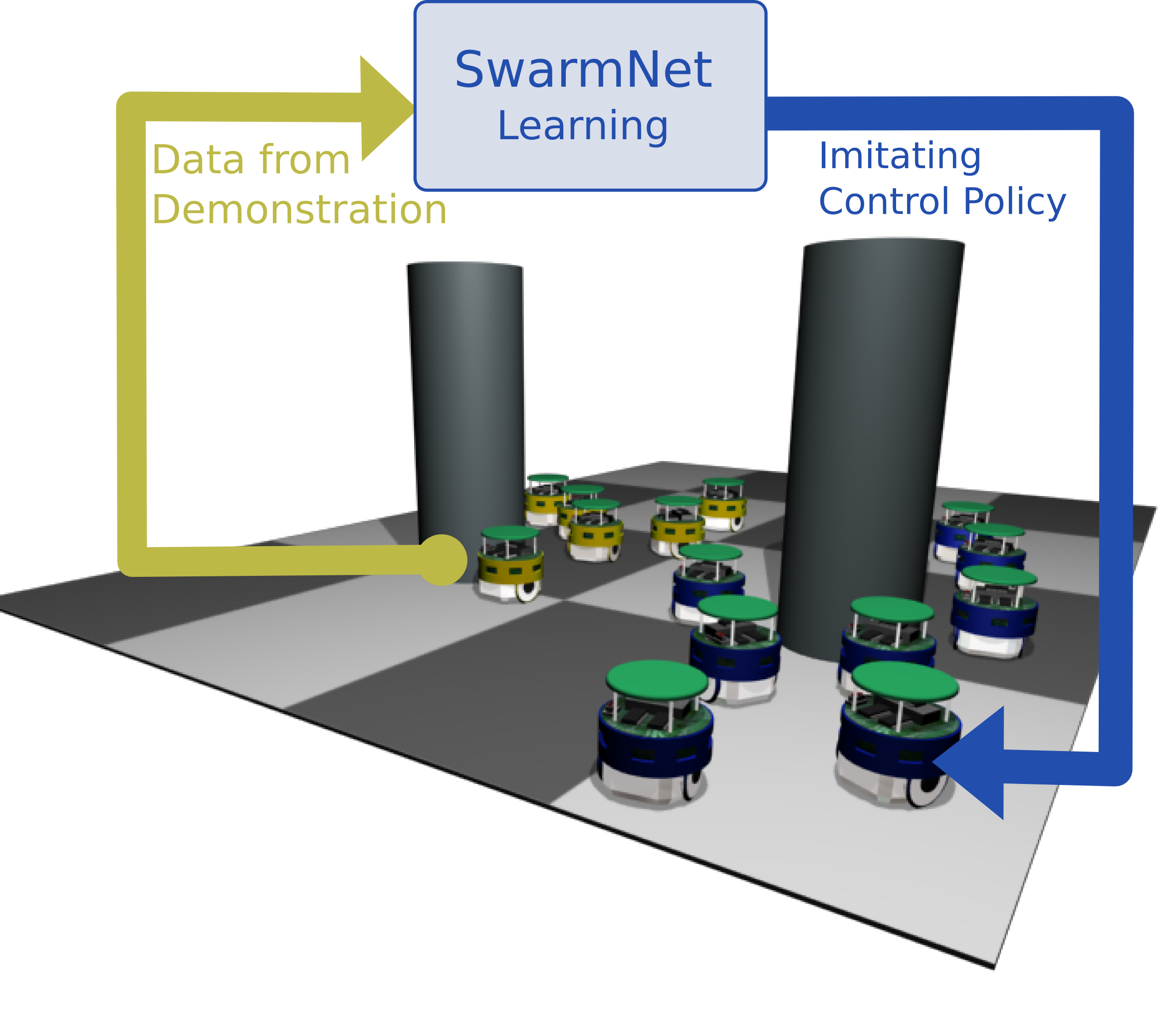}
    \caption{Visualization of the SwarmNet approach: data regarding the position of agents is recorded from an existing artificial or natural swarm (light colored robots). In turn, a graph neural network is learned that models the observed behavior. The trained SwarmNet can then be used as a policy to synthesize similar behavior for a multi-robot system (dark colored robots).} \label{fig:teaser}
\end{figure}

In this paper, we propose an alternative approach based on learning-from-demonstration (LfD)~\cite{schaal1999imitation} for specifying group behavior in multi-robot systems. The LfD methodology has a rich history in single-robot systems, with a number of successful applications in real-world tasks such as table-tennis, manipulation, locomotion, and helicopter flying~\cite{osa2018algorithmic}. We extend this methodology to the MRS case by introducing a novel graph neural network architecture that can be trained from execution traces of a swarm. Swarm systems can be naturally modeled as graphs, with nodes representing swarm members and edges describing interactions between these agents. The introduced graph neural network, called \emph{SwarmNet}, extracts all rules governing the group behavior from data alone, i.e., sequences of agent positions and velocities. Once a SwarmNet is extracted, it can be used to perform complex inferences including prediction, imitation or replacement of an existing swarm from a centralized view point (see Fig.~\ref{fig:teaser}).

\begin{figure*}[t!] 
    \centering
    \includegraphics[width=0.9\textwidth]{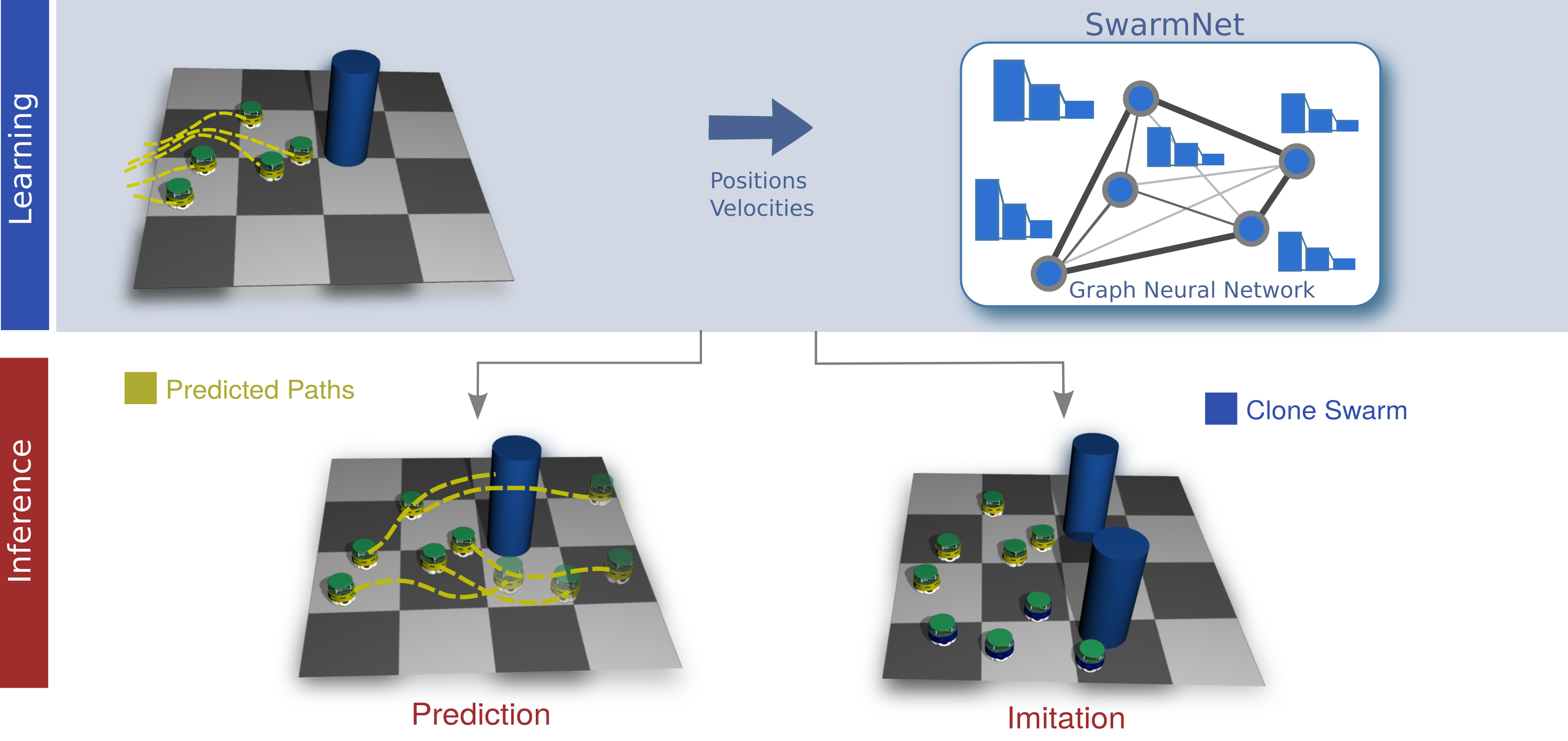}
    \caption{Overview of the proposed methodology: we collect training data from an observed swarm to learn a compact graph neural network representation called SwarmNet. In turn, it can be used to predict future behavior, augment an existing swarm with more agents, or create a clone of the swarm with similar behavior.}\label{fig:overview}
\end{figure*}

Moving beyond hand-coding the rules for inter-robot and robot-environment interactions, the approach presented in this paper allows for a data-driven methodology for the specification of swarm behavior. In particular, the contributions of this paper include: 

\begin{itemize}
\item  SwarmNet -- a graph neural network that can be trained from observations of a natural or artificial swarm. After training, SwarmNet can be used to (a) predict the behavior of all members of an observed swarm or (b) synthesize a \emph{clone swarm} -- a new swarm mimicking the trained behavior. 
\item Incorporation of contextual and environmental information, e.g., the location of obstacles and goals, into the learning and inference process of the swarm. As a result, a trained model of the swarm can be used as a reactive policy to control all agents. 
\item An extension of our approach, called SwarmNet$^{+}$ which can model nondeterministic swarms and environments. SwarmNet$^{+}$ captures the underlying probability distributions of group behavior. Sampling from this distribution generates  anticipated trajectories for all agents along with inherent uncertainties.
\end{itemize}

We evaluate our approach on a number of data sets generated from common models for flocking and swarming, e.g., the Boids model~\cite{Reynolds1999} and the Helbing model~\cite{Helbing2000}.

\begin{figure*}[ht!] 
    \includegraphics[width=\linewidth]{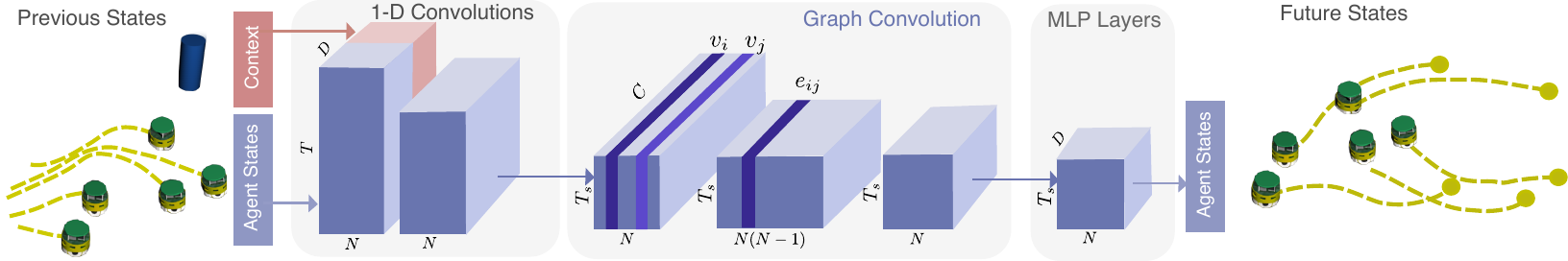}
    \caption{Diagram of the network architecture for SwarmNet: Letters along sides indicate dimensions. $T$ is the number of time steps, $N$ is the number of agents in the system, and $D$ is the length of the state vector. $H$ is the size of the encoded state vector, which is the size of the last layer of an MLP. Node states $v_i$ and $v_j$ are passed through Graph Convolution to produce interactions $e_{ij}$} \label{fig:SwarmNet}
\end{figure*}

\section{Related Work}
Research on modeling, prediction and control of multi-agent systems, e.g., teams of quadcopters, autonomous cars, or other forms of unmanned vehicles, has gained considerable attention in a variety of research disciplines. For an excellent recent survey of how such teams can be formalized, coordinated, and controlled, we refer the reader to~\cite{cortes2017coordinated}. Among the formalisms most widely used for representing and studying MRS and swarms are graph-theoretic~\cite{mesbahi2010graph} approaches and methods based on cost functions and auctioning processes~\cite{lagoudakis2004simple}. 
In contrast to the explicit modeling of MRS strategies presented in these approaches, one could also employ imitation and LfD to extract such policies~\cite{schaal1999imitation}. An early approach investigating the potential to copy the behavior of an MRS via imitation learning was presented in~\cite{chernova2007multiagent}. This approach used Gaussian mixture models to extract probability distributions underlying the agents' interactions. However, this approach was limited to discrete action spaces only. More recently, the work in ~\cite{zhan2018generative} used deep neural networks to produce generative models of multi-agent behavior. In contrast to graph theory based methodologies, the approach in~\cite{zhan2018generative} builds upon traditional, feedforward, and recurrent neural networks that do not explicitly model the team structure. Hence, the modelling and prediction task is reduced to a pure function approximation framework, which may neglect critical structural dependencies between the members of an MRS. This approach also assumes that macro-goals, e.g., a discrete set of sub-tasks an agent can take on, are available. In addition to imitation learning, modern reinforcement learning methods have also been used to generate MRS behavior~\cite{huttenrauch2017guided}. The approach in \cite{sartoretti2018primal} combines both reinforcement learning and imitation learning for extracting multi-agent navigation policies. The approach can deal with partially-observable domains, variable team-sizes, as well as complex environments and mazes. However, it also assumes a discrete set of actions and focuses on structured environments, such as warehouses and factories. In contrast, we focus on continuous action spaces and potentially unstructured nondeterministic environments. Further, our approach allows for the prediction of the most likely behavior of an observed swarm given previously seen behaviors. 

\section{Methodology}
In this section, we describe our methodology that is at the core of our approach. Fig.~\ref{fig:overview} depicts an overview of both the learning and inference process for MRS\footnote{We will henceforth use the term ``swarm'' and MRS interchangeably.}. Learning is achieved by recording execution traces of an observed swarm. Execution traces are discretely sampled trajectories specifying the position and velocity of each agent at time step $t$. Each trace represents a demonstration of the group behavior as observed in the swarm. The set of traces is, in turn, used to train \emph{SwarmNet} -- our novel graph neural network that encodes the dynamics among the agents of a swarm. Note that SwarmNet needs to account for both inter-agent interactions (e.g. how the members influence each other's behavior), as well as agent-environment interactions. Typically, swarms are also influenced by the current context and environmental variables. For example, in Fig.~\ref{fig:overview}, the obstacle needs to be taken into account to generate reasonable behaviors that avoid collisions with the environment. 

After training, the extracted SwarmNet can be used to implement two core functionalities, namely (a) prediction or (b) imitation of a swarm. In the prediction mode, we monitor the ongoing behavior of a swarm and predict the future locations of all involved agents. Such functionality is helpful, for example, in interdiction tasks~\cite{Hocaouglu19weapon, Kline18WTA, Guvenc18detection} in which a system has to generate recommendations on the least-cost and most-effective actions needed to thwart or disrupt the threat posed by an adversarial swarm. SwarmNet can also be used as a policy to generate a completely new swarm, which we refer to as \emph{clone swarms}. A clone swarm imitates the behavior seen during training. Since SwarmNet is trained in a data-driven fashion, a variety of input sources can be used to train models of multi-robot behavior, including previous swarming models, expert robot users, environmental and contextual attributes, and data from a biological swarm. 


\subsection{Network Architecture}

Fig.~\ref{fig:SwarmNet} describes the full architecture of SwarmNet. Input to the network is a set of previous robot states, as well as contextual information regarding the environment. Robot states are specified by a time window of positions and velocities of either (1) observed agents, or (2) controlled agents, depending on the mode of operation. Assuming an $N$-agent swarm, the dynamical state of agent $i \in \mathbb{N}$ can be written as $\boldsymbol{s}_i(t)= \left[\mathbf{x}_i(t), \mathbf{\dot x}_i(t)\right]$, where $\mathbf{x}_i(t)$ is the position and $\mathbf{\dot x}_i(t)$ is the velocity of agent $i$ at timestep $t$. Without loss of generality and for notational clarity, we will assume subsequently that agents live in a 2D space. The output of the network are the future state vectors $\mathbf{s}_i(t+1)$, given the state history of the swarm system and context. 

The states $\boldsymbol{s}_i(t)$ and context $\mathbf{c}(t)$ are concatenated and first passed through a series of one-dimensional convolution layers along the time axis. These convolutions allow the network to extract information regarding short-term dynamics of a single trajectory. In the Markovian case where the state of the next step only depends on the present step, the kernel of 1D convolutions would just assign a weight of 1 to the current state and 0s for all earlier steps. The evolution of the agents' dynamic states is a result of pair-wise interactions between them. The dynamics of the system and the underlying elementary interactions can be formulated as a graph, with nodes being the agents, their dynamical states being node states, the interaction relations being edges, and the interaction effects being edge states. To avoid any limiting assumptions or constraints, the states are embedded in a fully-connected directed graph which models all relationships, i.e., the learnable functions along the graph edges define whether agents are interacting or not. To process the interactions and update the node states for predicting the next step, a graph convolution over our representation is employed. The final set of operations in SwarmNet is a set of traditional fully-connected neural network layers. These final layers take the result of the graph convolution as input and generate a window of predictions over the future states of all agents. Subsequently, we will describe each one of these three sub-components in more detail. 



\subsection{1D Convolutions}

The set of one-dimensional convolutions in SwarmNet aims at incorporating a temporal context to the decision-making process. The approach used here avoids the usage of recurrent connections~\cite{hochreiter1997long}, which often introduces challenging nonlinearities thereby increasing the complexity of training and reducing the interpretability of the learned operations.

\begin{figure}[t!] 
    \centering
    \includegraphics[width=0.9\columnwidth]{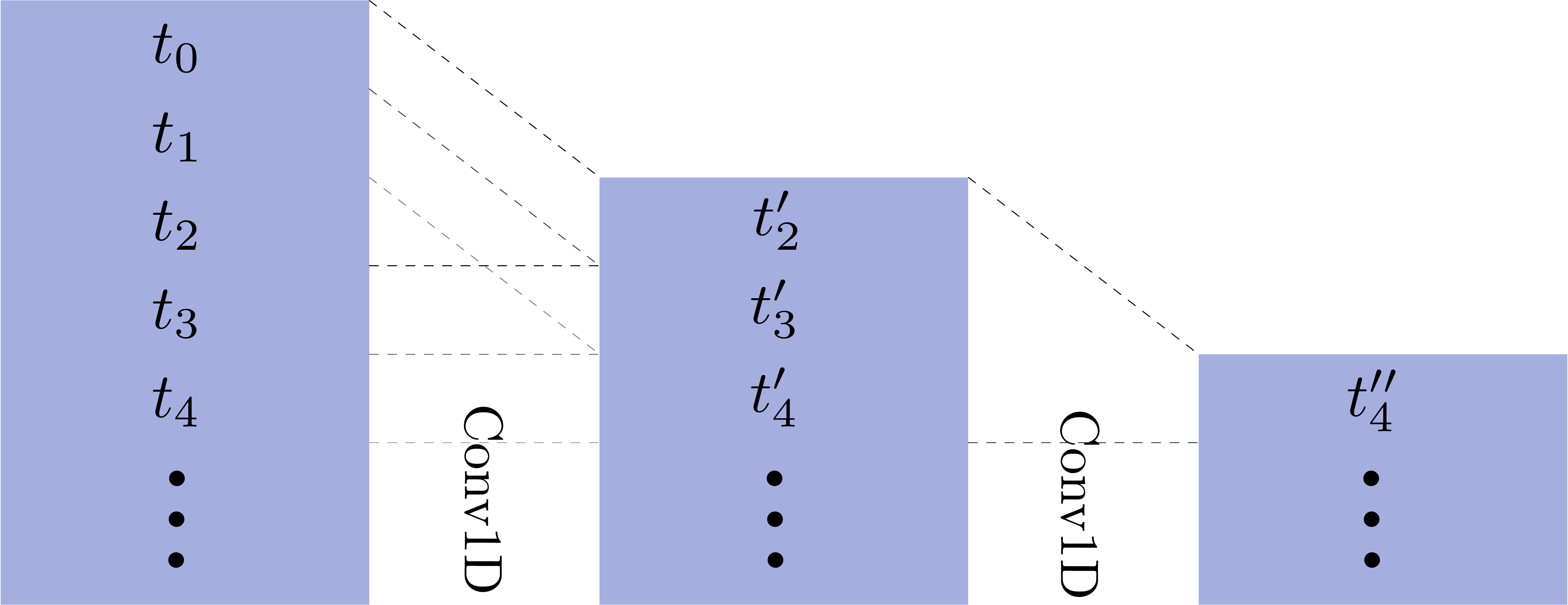}
    \caption{1D Convolution: for a kernel of size $K=3$, the time series is transformed by replacing each time step with the weighted sum of itself and its $2$ previous steps. Since the time series is not padded beyond step $0$, every convolution layer removes $2$ steps from the beginning. A filter with weights $(0, 0, 1)$ ignores earlier histories and only considers the present time step, thereby implementing a Markov assumption. A filter with weights $(0, -1, 1)$ effectively approximates the first order derivative, and one with $(1, -2, 1)$ approximates the second order derivative, etc. A group of such filters would be sufficient to capture the dynamics of histories.} \label{fig:Conv1D}
\end{figure}

In our approach, motion data is organized as a tensor (see Fig \ref{fig:SwarmNet}) with an overall dimension of $T \times N \times D$, where $T$ is the number of timesteps, $N$ is the number of agents, and $D$ is the dimension of the state vector, e.g., $\boldsymbol{s}_i(t) \in \mathbb{R}^4$ in 2D space. The time-series data is first passed through a series of 1D convolution layers along the time axis without padding, see Fig.~\ref{fig:Conv1D}. $L$ layers each with a kernel size of $K$ would condense a time-series of length $T_w=L(K-1)+1$ to a higher-order feature of length $1$. This can be considered an abstraction of the windowed temporal history into a more concise form used for prediction. A time-series of length $T$ will produce $T_{s}=T+1-T_w=T-L(K-1)$ such values after the kernel slides through every timestep in $T$. In particular, we chose $L=3$ layers with kernel size $K=3$, effectively having a window length $T_w=7$. Note, that similar to convolutional filters in image processing, we can use multiple 1D convolution filters in every layer. In our case, the different filters focus on different aspects of the trajectories and  identify temporal patterns that can be identified to generate better predictions (Fig.~\ref{fig:Conv1D}). We define $C$ to be the number of different 1D convolutional filters. Overall, the output of the entire 1D convolution step is condensed tensor of dimensionality $T_s \times N \times C$, which is then processed via a graph convolution.

\subsection{Graph Convolution}
The next step in SwarmNet is a graph convolution (GC) over the condensed state vectors. A GC is an operation applied uniformly across all nodes of a graph along with their local neighborhoods and is used to update the node and edge states of the next step, similar to the convolution in CNNs. As opposed to CNN, where the neighborhood of an element is the spatially neighboring elements inside an array, a matrix or a tensor, the neighborhood of a node is dictated by the connectivity of the underlying graph. 

\begin{figure}[ht!] 
    \centering
    \includegraphics[width=0.6\columnwidth]{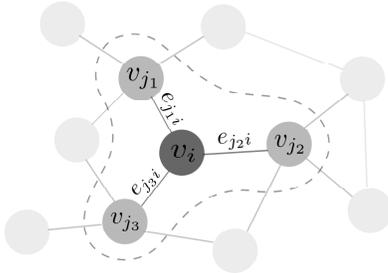}
    \caption{An example of graph convolution applied on the neighborhood of vertex $i$ which includes vertices $j_1$, $j_2$ and $j_3$. The dashed line encloses the neighborhood of vertex $i$. In a directed graph, edges sourcing from $i$ have no effect on $i$, and only edges targeting $i$ are considered. Graph convolution updates node attributes $v_i$ according to Eq. (\ref{eq:edge_update})-(\ref{eq:node_update})} \label{fig:GraphConv}
\end{figure}

The agent system is embedded in a directed graph $\mathcal{G}=(N,E)$, where $N$ is the set of nodes and $E=\{(i,j): i,j\in N, i \not = j\}$ is the set of edges. In $\mathcal{G}$, each node represents an agent and each of the $N(N-1)$ edges represent an interaction between a pair of distinct agents. The vector of attributes of node $i \in N$  is given by $\boldsymbol{v}_i \in \mathbb{R}^{d_v}$, where $d_v$ is the number of node attributes. Edges are uniquely identified by their connected nodes, thus we represent the attributes of the edge targeting node $j$ from $i$ (i.e., edge $(i,j) \in E$) by  $\boldsymbol{e}_{ij} \in \mathbb{R}^{d_e}$, where $d_e$ is the number of edge attributes. While node attributes model the state of the embedded agent, edge attributes model the interactive influence of source nodes onto target nodes. These interactions include pulling, pushing, and steering. For a thorough discussion of graph convolutions we refer the reader to ~\cite{Battaglia2018}.

\begin{figure*}[ht!] 
    \centering
    \includegraphics[width=0.85\textwidth]{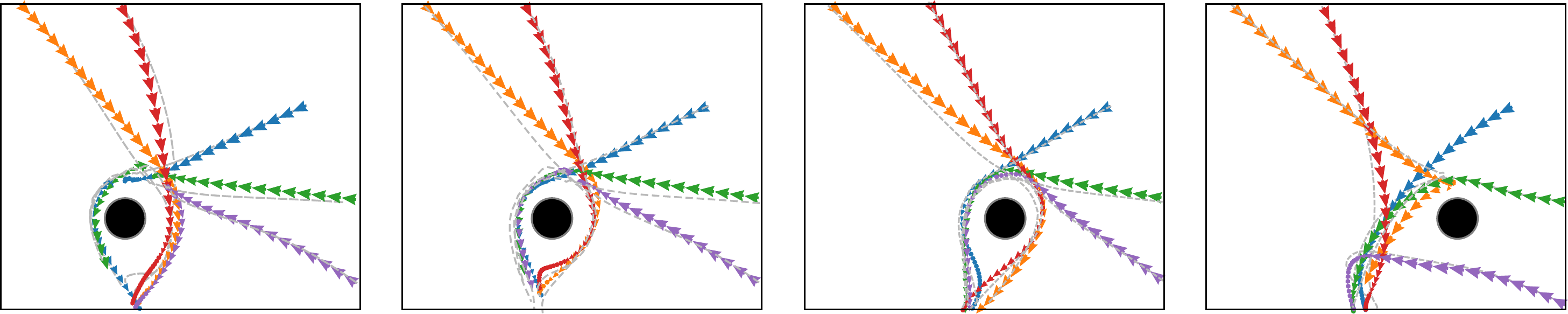}
    \caption{From left to right: four different predictions of the swarm behavior using SwamNet. In each one of the four experiments, the obstacle is place slightly more to the right. The colored arrows show the movement of 5 boids, while the gray dashed lines are the ground truth trajectories. The black circle represents the obstacle.\label{fig:movingObst}} 
\end{figure*}

Building upon the previous 1D convolution step, we use the condensed state vectors as the node states $\boldsymbol{v}_i$, see Fig. ~\ref{fig:GraphConv}. The node states for each member are determined by looking up the corresponding entries in the tensor which was generated from 1D convolutions, as can be seen in the graph convolution section Fig.~\ref{fig:SwarmNet}. Over time, agent states change due to the influence of interactions, which is in turn dependent on the new agent states. This alternating update of node states and edge states in the graph is propagated by GC. The GC process updates the node state as well as the edge state from step $t$ to step $t+1$ using the following functions:
\begin{equation} \label{eq:edge_update}
    \boldsymbol{e}_{ij}\leftarrow \phi^e(\boldsymbol{v}_i,  \boldsymbol{v}_j)
\end{equation}
\begin{equation} \label{eq:edge_aggr}
    \bar{\boldsymbol{e}}_i \leftarrow \psi^{\bar{e}} (\sum_{j \in \mathcal{N}_i}\boldsymbol{e}_{ji})
\end{equation}
\begin{equation} \label{eq:node_update}
    \boldsymbol{v}_i \leftarrow \phi^v(\boldsymbol{v}_i, \bar{\boldsymbol{e}}_i)
\end{equation}
where $i$ and $j$ are node labels. In Eq. (\ref{eq:edge_aggr}), $\mathcal{N}_i$ is the set of neighboring source nodes' states, so $\bar{\boldsymbol{e}}$ is an aggregation of the states of edges connected to node $i$.

The GC process involves three steps: 
\begin{enumerate}
\item Apply Eq. (\ref{eq:edge_update}) to all edges 
\item Aggregate states of connecting edges  using Eq. (\ref{eq:edge_aggr})
\item Apply Eq. (\ref{eq:node_update}) to update states of all nodes
\end{enumerate} 

Note that in Eq.(\ref{eq:edge_update}), we determine the edge state only by the present states of the connected nodes, and completely ignore the edge's past state. The GNN module treats the condensed time-series by 1D convolution as node states on an $N$-node graph. For each starting point of shape $N \times C$, each node's state is  sent to the edges attached to it for edge update according to Eq (\ref{eq:edge_update}). The edges effectively pull information from the two ends to deduce the interaction between them. We call this process \textit{node aggregation}. Following Eq.~(\ref{eq:node_update}), each node then aggregates the effect from incoming edges, resulting in a process called \textit{edge aggregation} along its previous state to compute the outcome. Further GNN layers would repeat the process until the ouput $\boldsymbol{v}_i(t+1)$ of the last layer is taken as the prediction of dynamical states, i.e. positions and velocities, of the agents. In the proposed SwarmNet architecture, we perform only a single such GC operation as described above. The functions $\phi^e$, $\phi^v$ and $\psi^{\bar{e}}$ are approximated using traditional multi-layer perceptrons (MLP), i.e., feed forward neural networks. The MLPs apply on the last dimension. A final MLP after edge aggregation transforms the tensor to one with shape $T_s\times N\times D$, giving each of the $T_s$ starting points the prediction of their next steps. 


\subsection{Loss Function}

Training of the network is performed in a supervised fashion by comparing the predicted agent states (containing positions and velocities) with the ground-truth states in the time series. The mean squared-error (MSE) is used as the loss function $L$ for training and as the metric for evaluation, with
\begin{equation}
    L = \frac{1}{2DNT_{s}}\sum_{t=1}^{T_{s}} \sum_{i=1}^{N} \left(\boldsymbol{s}_i(t) - \boldsymbol{s}^{*}_i(t)\right)^2
\end{equation}
where $\boldsymbol{s}^{*}_i(t)$ is the ground-truth state vector of agent $i$ at step $t$. The loss is normalized over the "natural skip", i.e., the MSE of state vectors between two consecutive steps in the ground truth trajectories, $\bar{L}$. 
\begin{equation}
    \bar{L} = \frac{1}{2DN(T-1)} \sum_{t=1}^{T-1}\sum_{i=1}^N \left(\boldsymbol{s}^{*}_i(t+1) - \boldsymbol{s}^{*}_i(t)\right)^2
\end{equation}
Given the above normalization factor, the normalized loss can be calculated using the fraction $L_{norm} = \frac{L}{\bar{L}}$. Normalization rectifies the dependence of prediction error on the intrinsic spacing of the ground-truth trajectories. 


\subsection{Multistep Predictions and Curriculum Learning}
\label{sec:curriculum}

To enable multistep prediction, the shifting windows of the time-series leading to the $T_s$ starting points are stacked such that the restructured time-series has the shape $T_{s} \times T_w \times N \times D$, where $T_{s}=T-L(K-1)$ and $T_w=L(K-1)+1$ as discussed earlier. Now that 1D convolution acts along $T_w$, eventually the prediction of next steps has the shape $ T_{s} \times 1 \times N \times D$ and is appended to the input along the $T_s$ dimension. Then, with the first state dropped, the time-window is shifted by one-step and is the temporal history of states for the prediction of the next step. This procedure can be iteratively applied to create prediction horizons of arbitrary length.

We scaffold the learning process by training with an increasing horizon for the predictions. More specifically, we increase the required number of prediction steps gradually through the course of training, from 1 to 10 steps. This \emph{curriculum learning} scheme has the benefit of starting with a simpler version of the task (i.e. using limited prediction horizon) and slowly exposing the network to more complex, long-term predictions.  The approach also encourages the model to learn the true dynamics of the system than can be unfolded for arbitrary time steps, rather than potentially overfitting to fixed-term predictions.

\subsection{Uncertainty, Noise and Nondeterminism}

Swarms and multi-robot systems are typically acting in nondeterministic environments in which perceptual data is noisy and the effect of actions uncertain. We extend our methodology to nondeterministic environments by leveraging recent theoretical insights connected to \emph{Bayesian} deep learning. In particular, the work in \cite{Gal2016Uncertainty} shows that estimates of model uncertainty can be generated from a neural network using the Dropout~\cite{Srivastava2014} algorithm. Dropout is a training algorithm in which connections of a neural network are randomly activated and disactivated. This is achieved using a dropout probability $p$ which describes the probability of an input activation being dropped. After training a network with Dropout, we can use the same technique to generate different outputs for the same input, i.e., disactivate layer input with probablity $p$. In such a case, each forward pass through the network is called a \emph{stochastic forward pass} and can be seen as a sample from the underlying probability distribution. According to \cite{Gal2016Uncertainty}, such stochastic forward passes in a deep network will be an approximation of variational inference in a Gaussian process. In our case, the set of stochastic forward passes  ${\cal S} =\{\boldsymbol{{\hat s}}_i^{1}(t+1), ..., { \boldsymbol{\hat{s}}}_i^{S}(t+1) \}$ represents $S$ samples from the probability distribution over the future states of agent $i$, i.e., the distribution defining the range of different states the agent can be as a result of nondeterminism. Recursively sampling for longer horizons of the future generates the anticipated trajectories for all agents along with inherent uncertainties. In the remainder of this paper, we will refer to versions of our network that use this approach as \emph{SwarmNet}$^{+}$. It is important to note, that SwarmNet$^{+}$ can  generate multiple different, potentially conflicting or bifurcating predictions for the same input state to the network -- a property that is typically not possible in standard neural network training approaches.    


\section{Experiments and Results}
We generated training data using popular techniques for synthesizing swarm and MRS behavior, e.g., flocking and swarming models. In turn, we compared different versions of SwarmNet to the graph neural network (GNN) in \cite{Kipf2018} and the LSTM network method in \cite{hochreiter1997long}. Since SwarmNet shares strong similarities with GNN approaches such as \cite{Kipf2018}, we tested different ablations and additions to \cite{Kipf2018} based on the insights in this paper. In addition, we performed a variety of experiments to investigate sample-efficiency and other critical aspects of SwarmNet.

\begin{table*}[t!] 
    \centering
    \begin{tabular}[c]{l | c c c | c c c}
        \hline \hline
         & \multicolumn{3}{c|}{\textbf{5 Steps}}  & \multicolumn{3}{c}{\textbf{40 Steps}}  \\
        \cellcolor{gray!15} Method &         \cellcolor{gray!15} Boids &         \cellcolor{gray!15} Helbing &         \cellcolor{gray!15} Chaser &         \cellcolor{gray!15} Boids &         \cellcolor{gray!15} Helbing &         \cellcolor{gray!15}Chaser\\ 
        \hline
        Kipf's GNN & $0.4288\pm0.0182$ & $0.4374\pm0.0179$ & $0.1391\pm0.0006$ & $17.45\pm1.00$ & $19.61\pm0.43$ & $14.23\pm0.05$ \\ 
        
        LSTM & $0.8992\pm0.0098$ & $1.2241\pm0.0098$ & $0.2013\pm0.0020$ & $41.20\pm0.24$ & $42.88\pm0.75$ & $126.3\pm3.5$ \\

        SwarmNet & $0.2813\pm0.0012$ & $0.1000\pm0.0050$ & $\mathbf{0.0152\pm0.0002}$ & $10.47\pm0.91$ & $3.855\pm0.305$ & $\mathbf{3.691\pm0.042}$ \\
        
        SwarmNet (Context) &  $\mathbf{0.2338\pm0.0024}$ &   $\mathbf{0.0317\pm0.0019}$ & $\mathbf{0.0152\pm0.0002}$ &   $\mathbf{2.778\pm0.066}$ & $\mathbf{0.484\pm0.023}$ & $\mathbf{3.691\pm0.042}$ \\
        
        \hline\hline
    \end{tabular}
\caption{Normalized MSE loss for short term prediction (left, 5 steps) and long term prediction (right, 40 steps)} \label{Table: MSE_loss}
\end{table*}

\begin{figure}[t!]
    \centering
    \includegraphics[width=0.85\columnwidth]{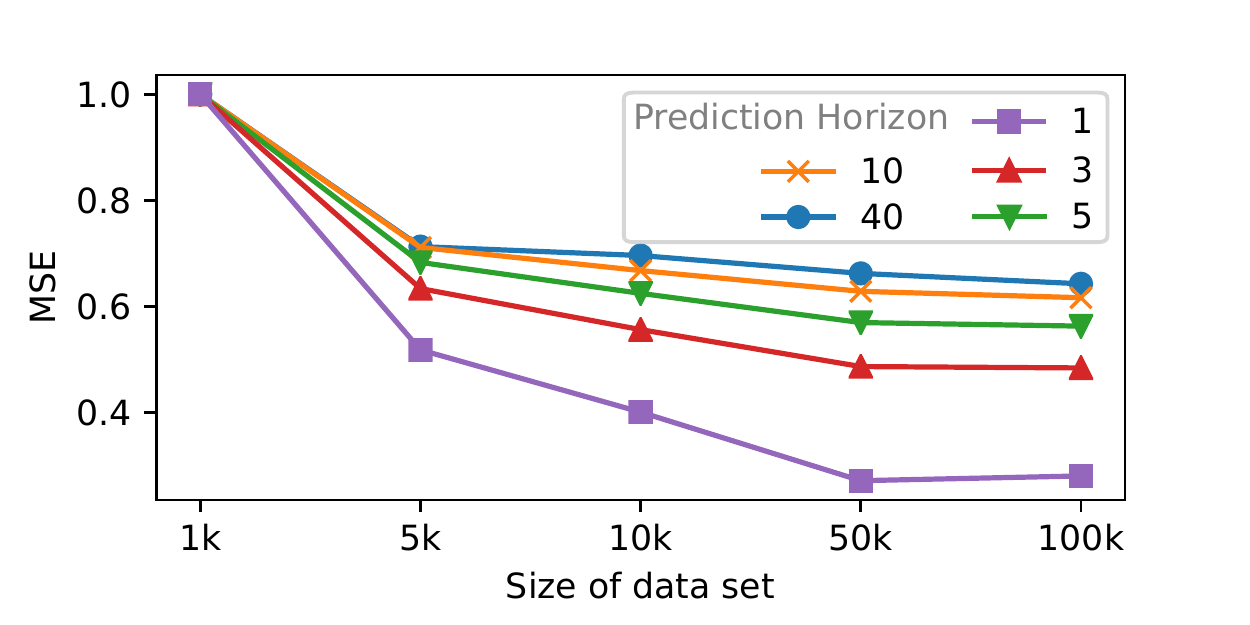}
    \vspace{-0.3cm}
    \caption{Prediction error for different horizon lengths and sizes of training set with boids.\label{fig:sampleSize}}
\end{figure}

\subsection{Swarm Data Sets}

Our data set consists of motion data produced using the Boids model~\cite{Reynolds1999}, the Helbing model~\cite{Helbing2000}, and a simple chaser model. In the Boids model, $N=5$ agents demonstrate flocking behaviors while approaching a common goal location and avoiding obstacles. The Helbing model~\cite{Helbing2000} generates swarm behavior via repulsive forces only. The final model, called chaser, places a set of agents on a circle. Each of these agents generates steering actions that make it chase another agent of the swarm. For each of the above models we generate data sets by running the simulation for about 50 time steps while recording the agent positions and velocities, as well as the environmental variables. In both the Boids and Helbing model, the environmental variable is the positions of obstacles. 


\subsection{Prediction Accuracy}
The first experiment focuses on the prediction accuracy when compared to other methods, in particular the method in~\cite{Kipf2018} and LSTMs~\cite{hochreiter1997long}. Table~\ref{Table: MSE_loss} summarizes the loss of the methods across the three data sets. We performed the experiment for prediction horizons of 5 steps and 40 steps. All methods were trained with 50K demonstrations of the corresponding swarm behavior. The best performance is achieved by SwarmNet with contextual inputs. Removing the context information has a significant impact on the SwarmNet performance, in particular in the long-term prediction case (right side of the table). On the Boids data set, the error for 40 step predictions jumps from $2.778$ to about $10.47$ when contextual information is omitted. In general, SwarmNet outperforms all other methods on all data sets and in both conditions. An interesting aspect of these results is that SwarmNet was only trained on data for predictions of up to 10 steps, as previously described in Sec.~\ref{sec:curriculum}. Despite that, it correctly learned to make accurate predictions for 40 steps and beyond indicating that it focused on the true dynamics of the task. 

Fig.~\ref{fig:movingObst} shows the prediction results for different locations of the obstacle (black circle). We see that the SwarmNet predictions are qualitatively implementing the correct swarming and avoidance behavior, even if small deviations from the ground truth occur. The rightmost example in Fig.~\ref{fig:movingObst} shows an interesting behavior in which the orange agent is first trying to circumvent the object on the right side and then turns to head back to goal location. While unintuitive, this behavior is also existent in the ground truth data.

\begin{figure}[t!] 
    \centering
    \includegraphics[width=\columnwidth]{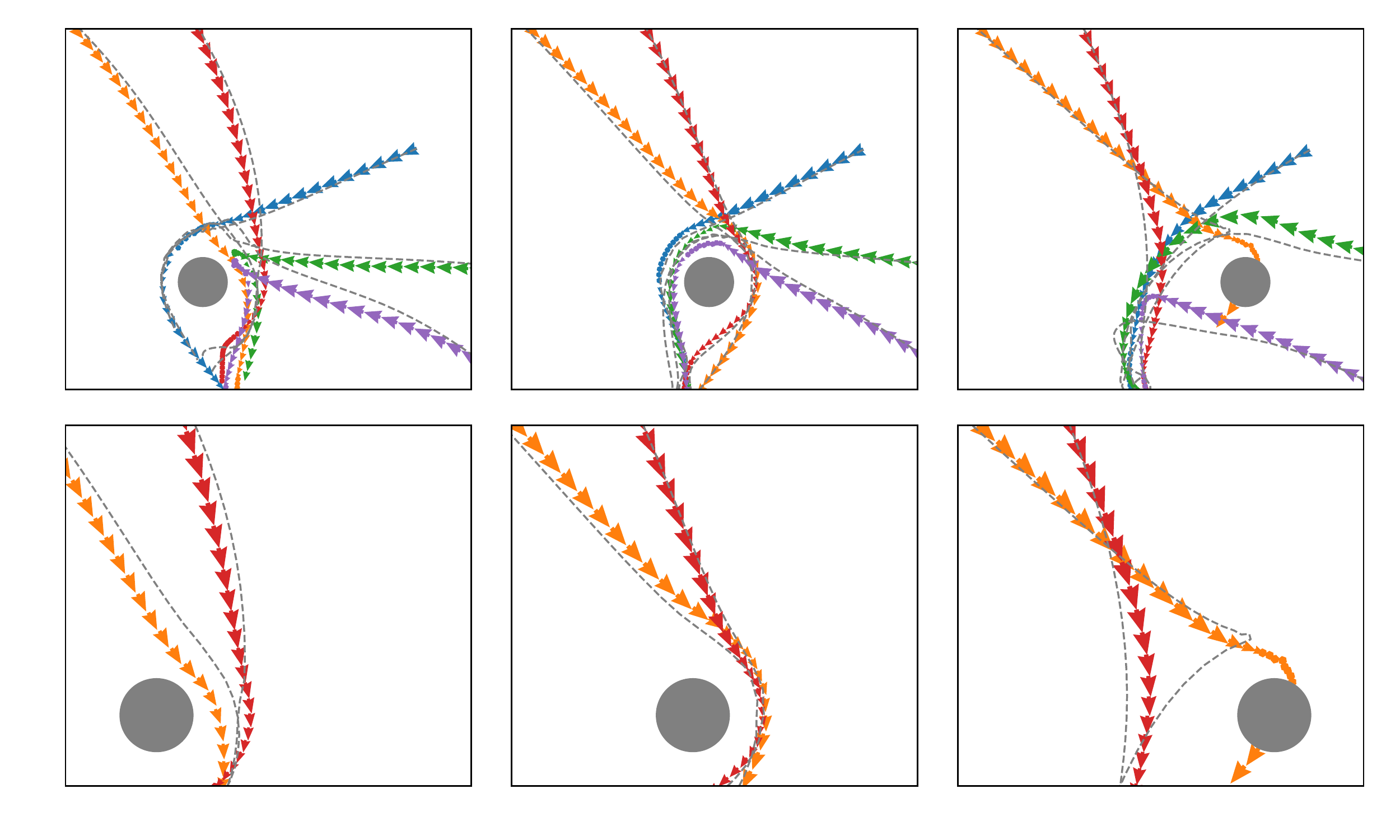}
    \caption{Predictions for the behavior of a swarm generated from a network trained on 1000 demonstrations of boids simulation. The top row shows the movements of all agents. The bottom row highlights (for visibility) the movements of only two agents. \label{fig:MovingObstacle_1k}}
\end{figure}

Next, we investigated the influence of the \emph{training set size} and the \emph{prediction horizon} on the prediction results. Fig.~\ref{fig:sampleSize} shows the MSE for five different horizon lengths (1-40 steps into the future) trained with data sets ranging from 100K demonstrations down to only 1K demonstrations. It is interesting to note that, in the case of long-term prediction, only when trained with less than 5K demonstrations does the MSE deteriorate. Even in the case of training on only 1K samples, SwarmNet still generates (qualitatively) reasonable swarm behavior that executes the intended task. Fig.~\ref{fig:MovingObstacle_1k} shows the behavior of a network trained on 1K examples of the Boids task. The generated predictions still follow the trend of the ground truth data, with one exception in which an agent trajectory (orange) is predicted to go through the obstacle.   


\begin{figure*}[t!]
    \centering
    \includegraphics[width=0.24\linewidth]{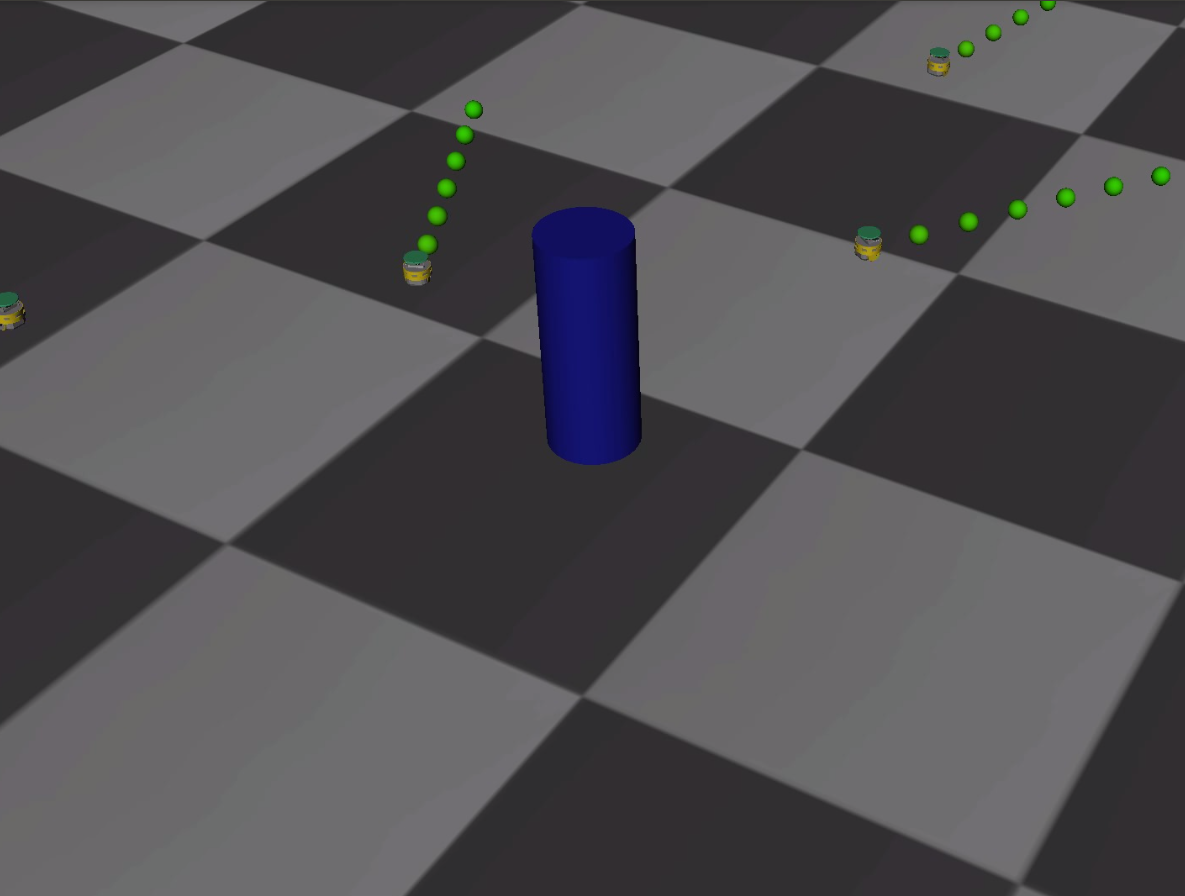}
    \includegraphics[width=0.24\linewidth]{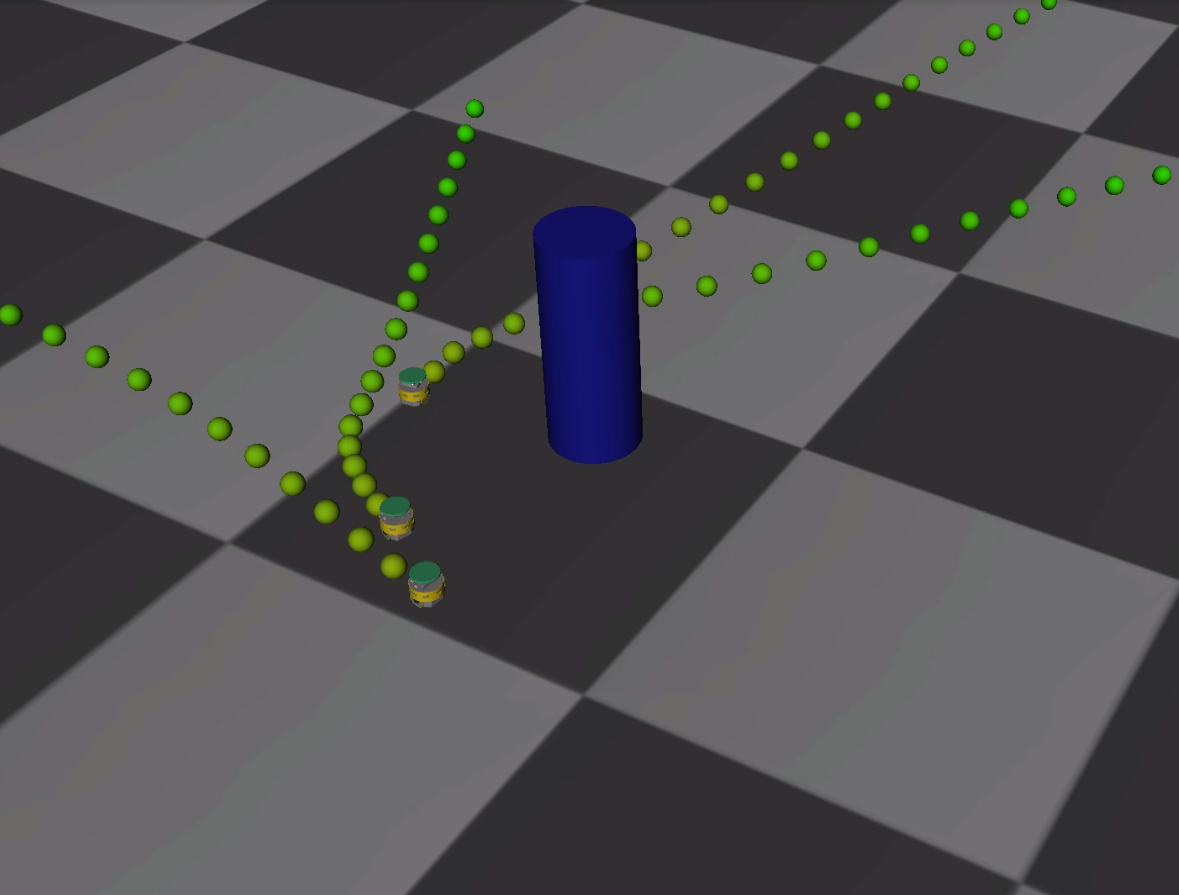}
    \includegraphics[width=0.24\linewidth]{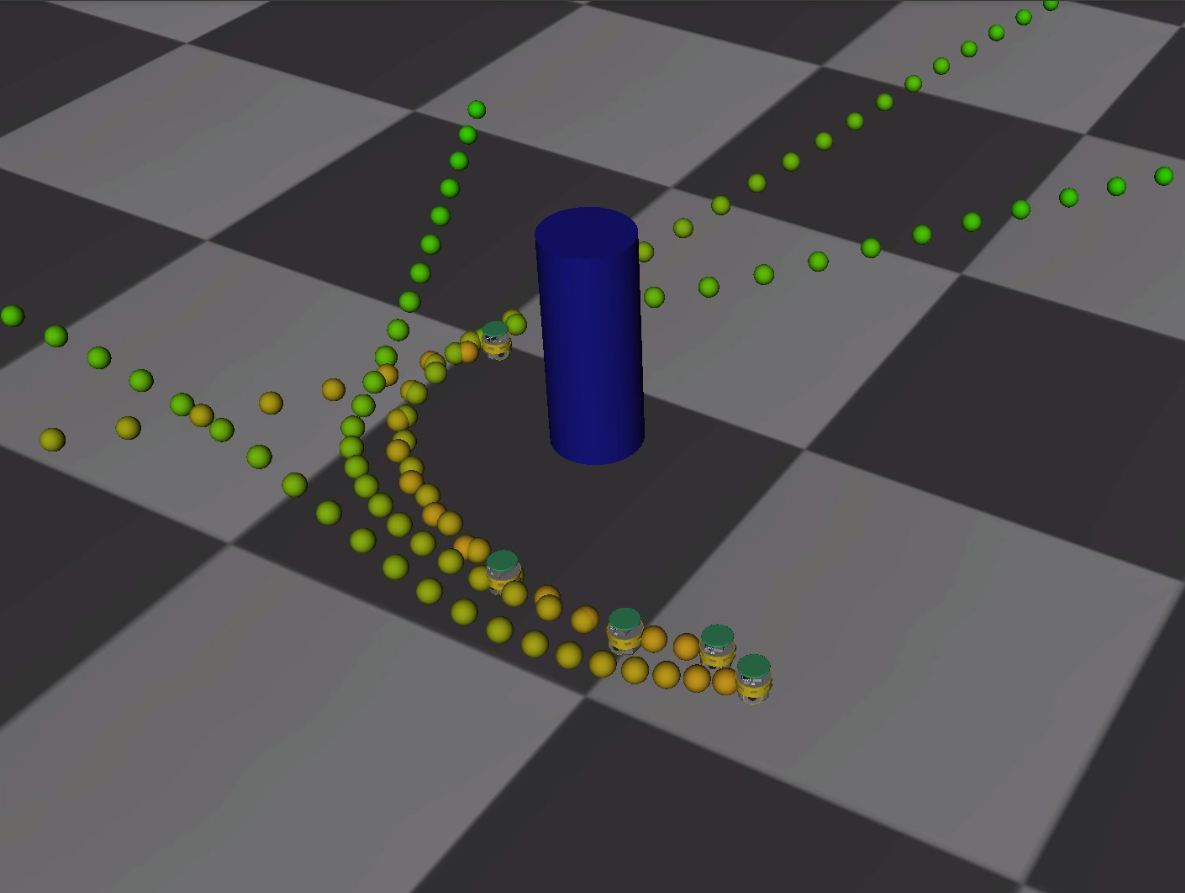}
    \includegraphics[width=0.24\linewidth]{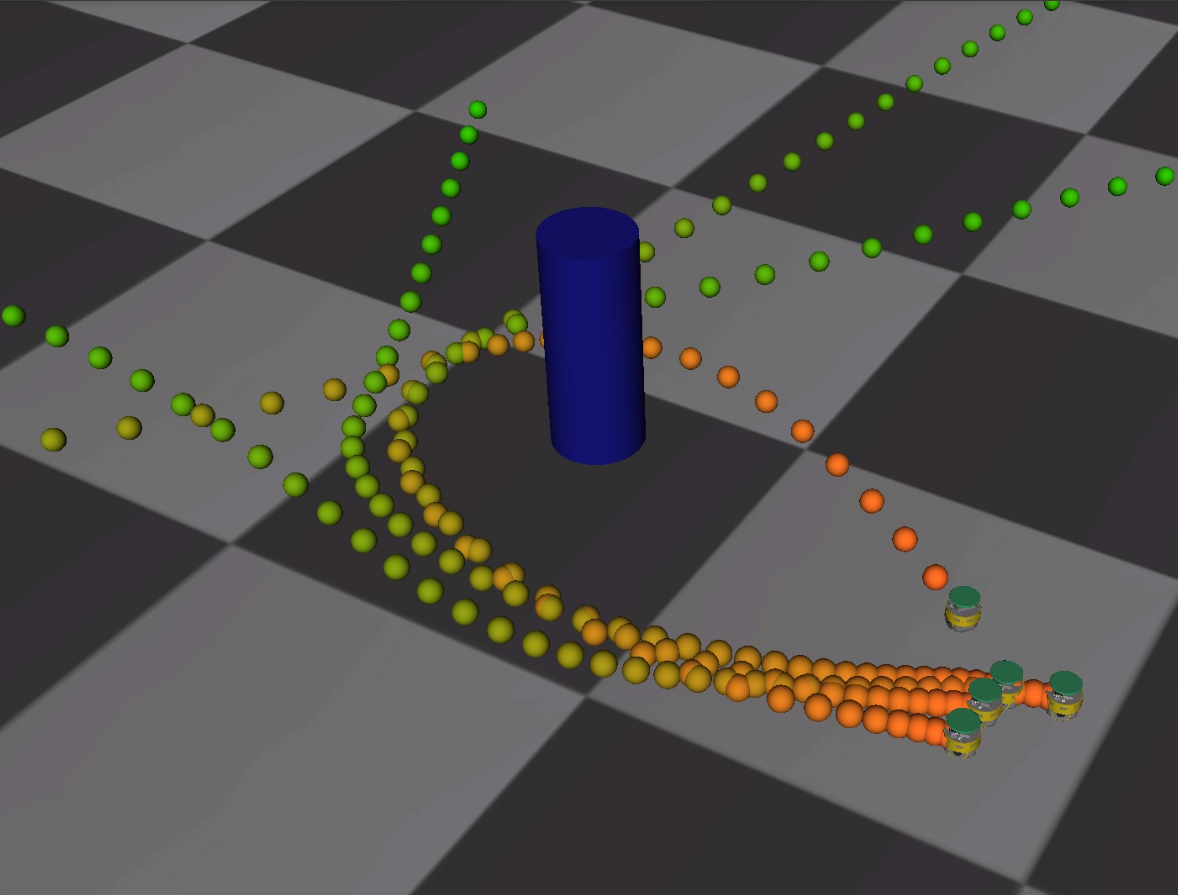}\vspace{0.3cm}
    \includegraphics[height=2.6cm]{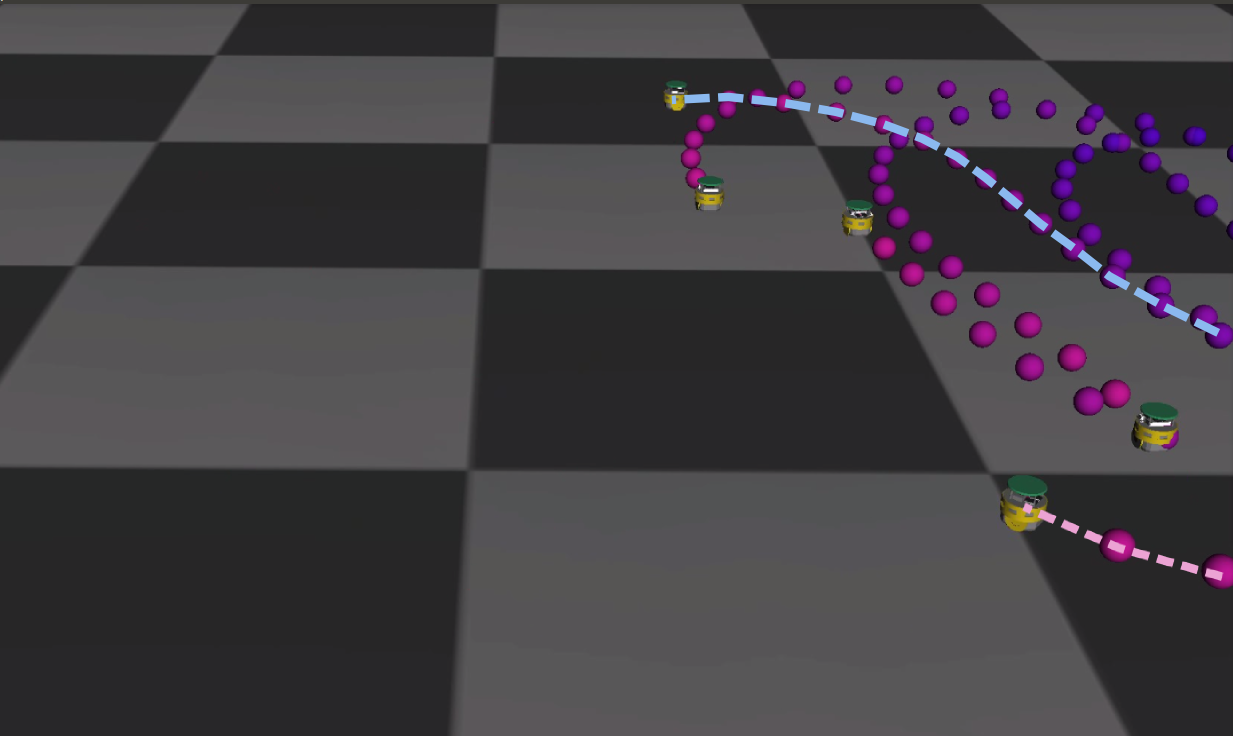}
    \includegraphics[height=2.6cm]{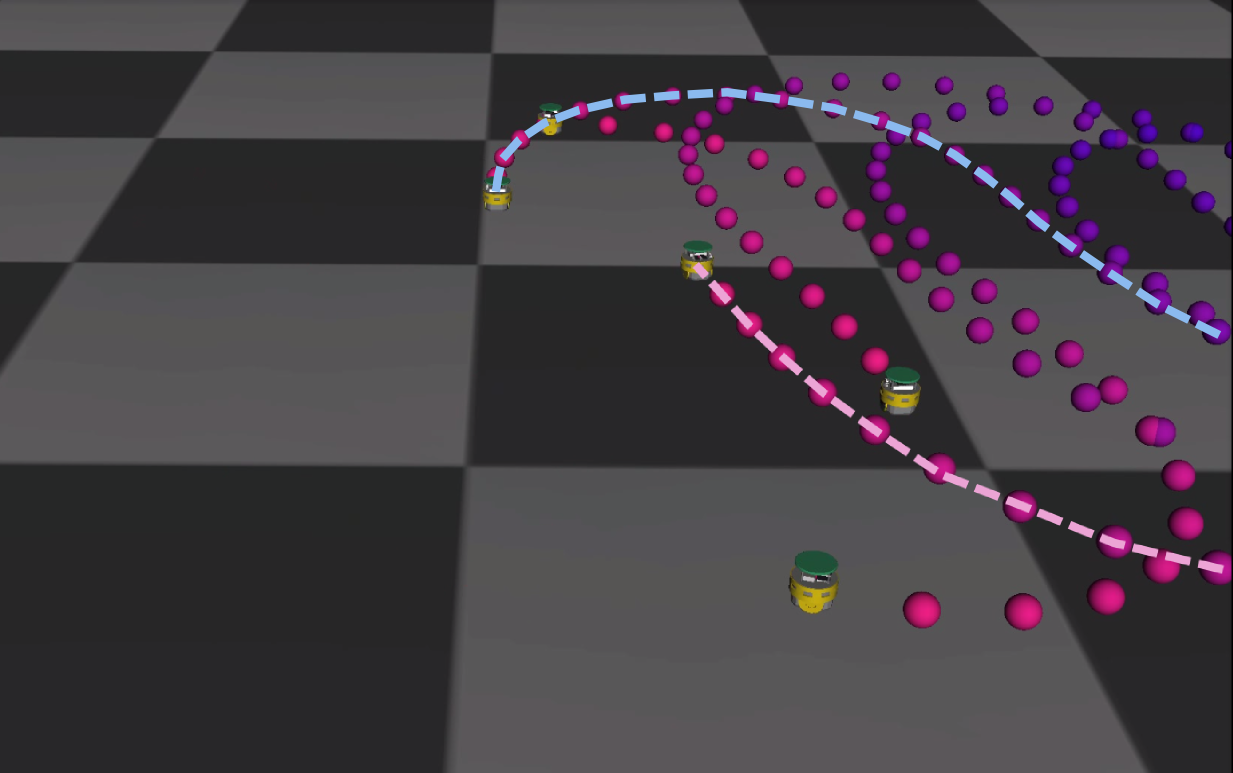}
    \includegraphics[height=2.6cm]{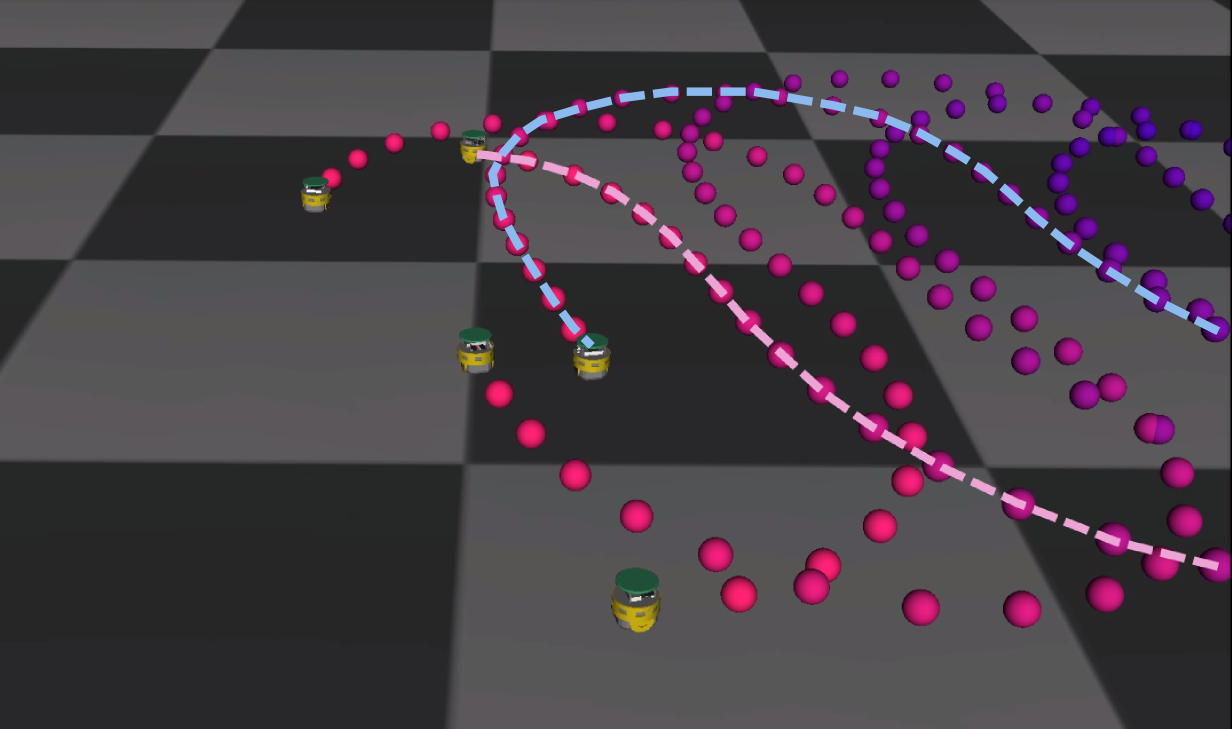}
    \includegraphics[height=2.6cm]{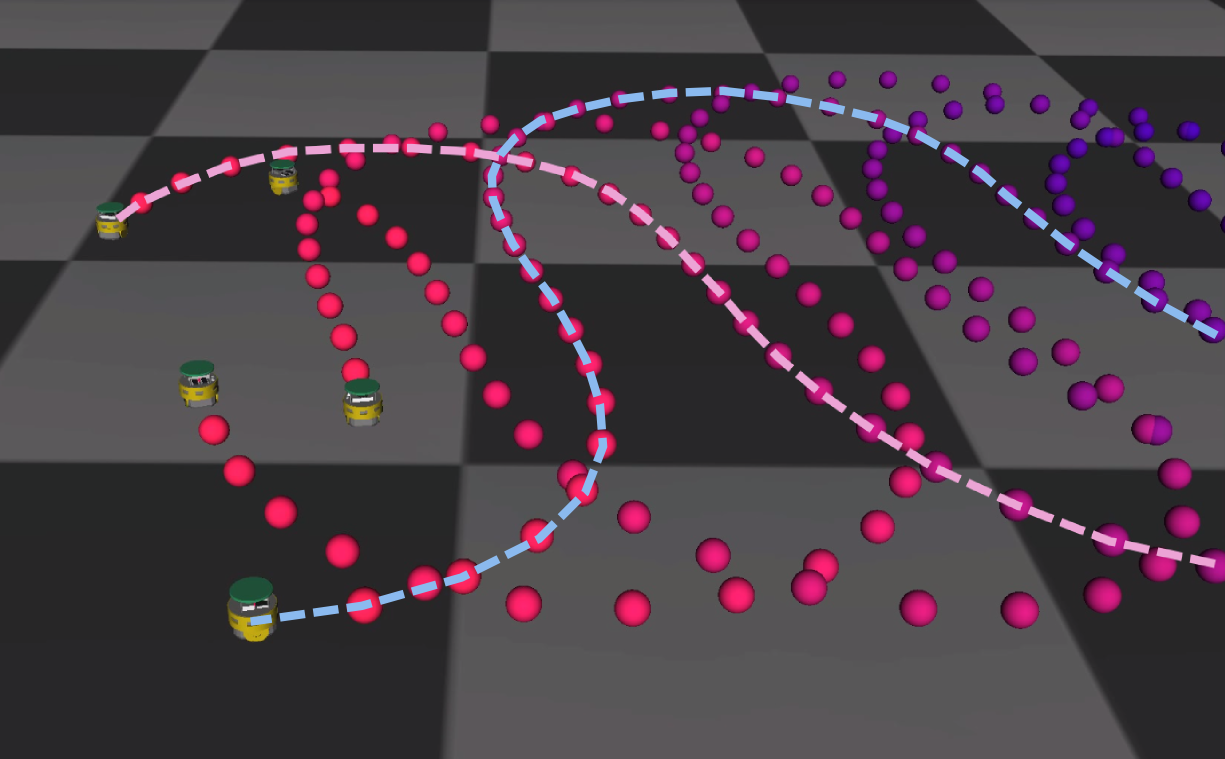}
    \caption{Top sequence: Image sequence from the robot simulation of an ePuck MRS. All robots are controlled by a SwarmNet which was trained on the Boids model. The colored dots represent the states of the agents at different moments in time. Bottom sequence: A SwarmNet control policy implementing the chase behavior. \label{fig:sequence1}}
\end{figure*}

\subsection{Comparison to the Kipf Model}
\begin{figure}[ht!] 
    \centering
    \includegraphics[width=\linewidth]{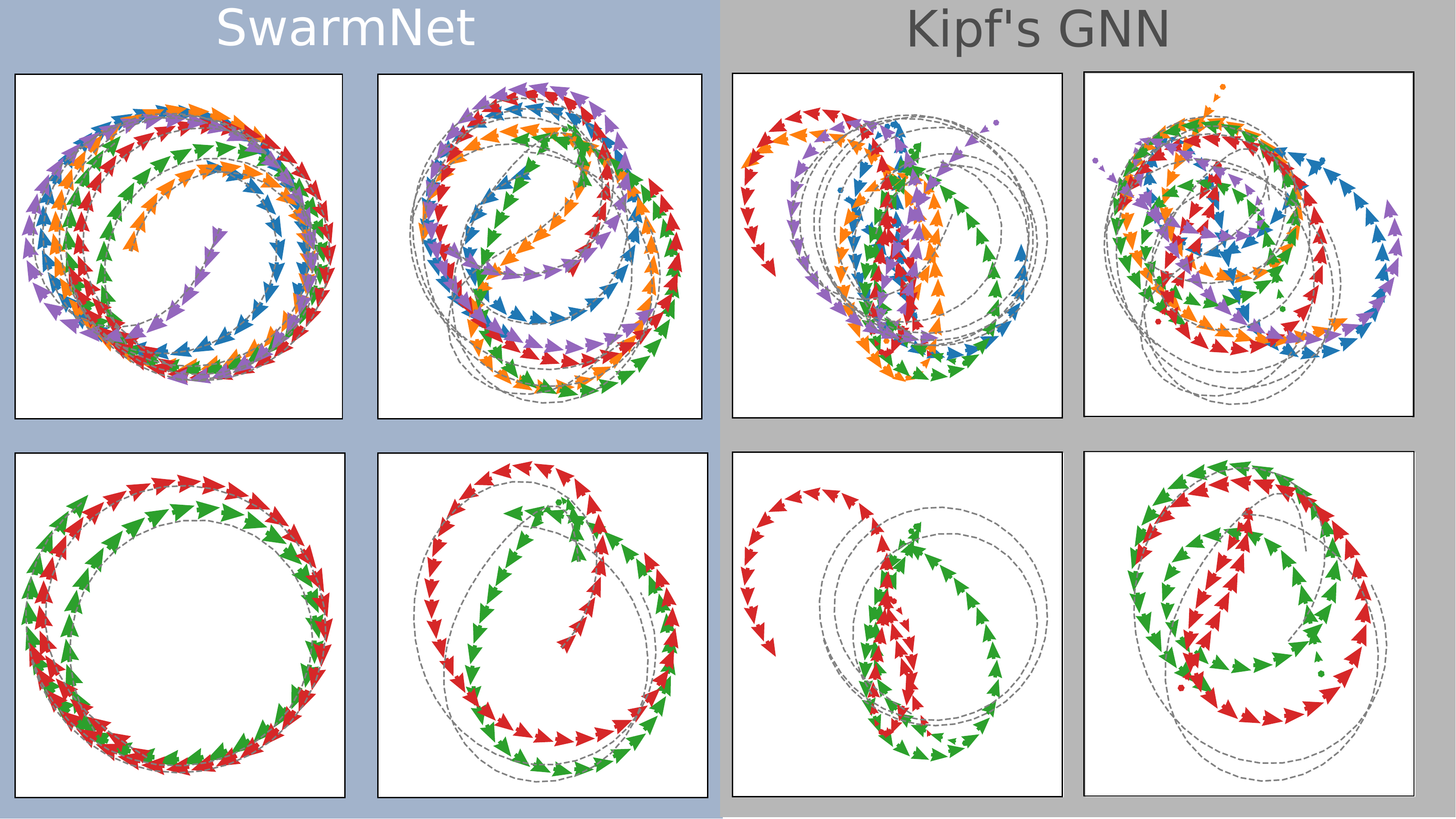}
    \caption{Qualitative comparison of SwarmNet and the approach in \cite{Kipf2018}. Two samples of predicted trajectories (scattered arrows) against the ground truth trajectories (grey dashed lines) for a 5 agent system performing chasing motions. Top row: traces of all 5 agents. Bottom: traces of only two agents for visibility. Only the starting $T_{seg}$ states are provided to the model, and the prediction is done consecutively to the end. All graphs use data from test set.\label{trajectories}}
\end{figure}

SwarmNet builds upon recent developments and advances in the field of graph neural networks~\cite{Battaglia2018}. In particular, it shares similarities with the method proposed by Kipf~\cite{Kipf2018}. However, our method incorporates new components, in particular the 1D convolutions, use of context, and curriculum-based training. To better understand the effects of the individual components of our network, we make changes (ablations and additions) to the original Kipf method to see how they affect performance. First, we noted that SwarmNet can be seen as a variant of only the decoder part of Kipf's approach. Hence, we used the decoder as a starting point and then performed repeated experiments in which we added (a) MLP layers for edge aggregation and context (b) 1D convolutions. Tab.~\ref{Table:ablation}, shows the results of this comparison. We can see that the decoder of Kipf's model, when combined with both 1D convolutions and context variables, yields MSE error values comparable to our results. The different neural network architectures shown in Tab.~\ref{Table:ablation} can be seen as a (discrete) spectrum that shows the effects of gradually transitioning from the original Kipf GNN to our SwarmNet model. Our model still slightly outpaces Kipf's decoder augmented with Conv1D and context. This last difference is due to the curriculum learning approach to training. For a qualitative comparison between SwarmNet predictions and the Kipf GNN predictions, see Fig.~\ref{trajectories}.

\begin{table}[h!]
    \centering
    \begin{tabular}[c] {l | c | c}
    \hline\hline
    \cellcolor{gray!15}Method & \cellcolor{gray!15}5 Steps & \cellcolor{gray!15}40 Steps \\
    \hline
    Kipf's GNN & $0.4288\pm0.0182$
  & $17.45\pm1.00$ \\
    Decoder & $0.2654\pm0.0070$ & $3.639\pm0.075$ \\
    Decoder (Context) & $0.2717\pm0.0034$ & $3.873\pm0.090$ \\
    Decoder+Conv1D & $0.2584\pm0.0040$ & $3.916\pm0.388$ \\
    Decoder+Conv1D (Context) & $0.2491\pm0.0018$ & $3.101\pm0.046$\\
    SwarmNet (Context) &  $\mathbf{0.2338\pm0.0024}$ & $\mathbf{2.778\pm0.066}$ \\
    \hline\hline
    \end{tabular}
\caption{Comparison of MSE error on Boids data of SwarmNet, the original Kipf model, as well as different new models that are created from ablations and additions to the decoder part of the network in \cite{Kipf2018}.} \label{Table:ablation}
\end{table}

\subsection{SwarmNet as a Policy for Controlling Clone Swarms}
As introduced in Sec.~\ref{sec:intro}, a trained SwarmNet can also be used to control a new swarm. The objective in this case would be to generate new behavior that is similar to that observed in the training set. To test this property, we used the output velocities generated by SwarmNet as control signals for a simulated swarm of robots. In particular, we are using a simulation of the ePuck robot. 
Fig.~\ref{fig:sequence1}  shows the result of using a SwarmNet, trained on Boids data and the Chaser data respectively, as a control policy for the ePuck MRS. In both cases, generated behavior of the clone swarm reproduces the dynamics inherent to the training data as discussed before.

\begin{figure}[ht!] 
    \centering
    \includegraphics[width=\columnwidth]{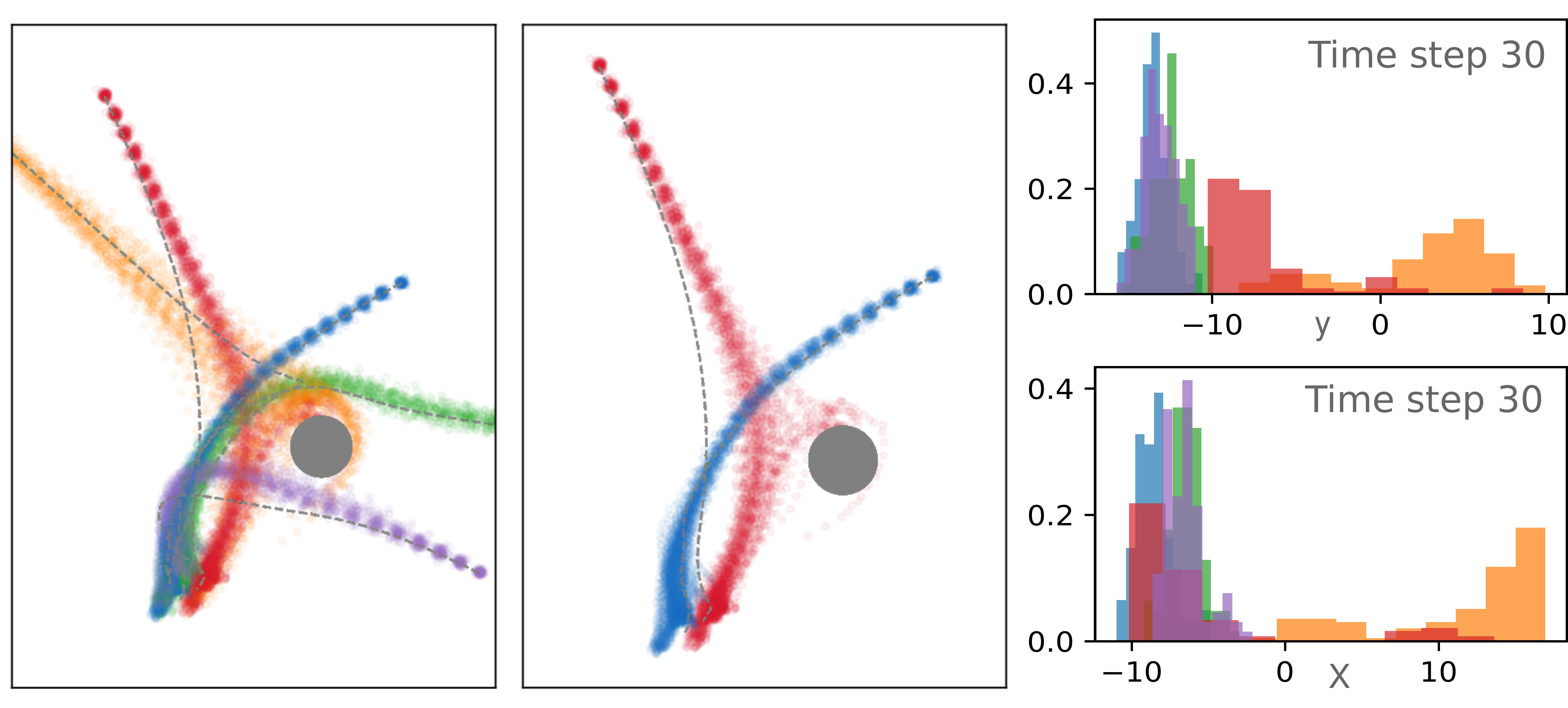}
    \caption{Predictions for the \emph{nondeterministic} behavior of a swarm with 5 boids in the presence of uncertainty. Left graph: the stochastic output predictions of the SwarmNet$^{+}$ network for all agents. Middle: the predictions for only the blue and the red agent, for visual clarity. Right: The probability distributions over x and y coordinates of all agents at time step 30. We can see that the predictions for the red agent bifurcate to the left and right side of the obstacle. \label{fig:stochastic}}
\end{figure}

\subsection{Uncertainty}

Finally, we also investigated how well the SwarmNet$^{+}$ extension can deal with uncertainty, noise and nondeterminism. In particular, we simulated both perceptual noise, as well as actuation noise. Perceptual noise was incorporated by adding a value sampled from a univariate normal distribution ${\cal N}(0,1)$ to input for each dimension separately. Actuation noise is simulated by randomly dropping out neurons in test time. Fig.~\ref{fig:stochastic} depicts the stochastic outputs of our SwarmNet$^{+}$ model for the Boids data. We can see in Fig.~\ref{fig:stochastic} (left) that the predictions now form envelopes according to the uncertainty at different time steps in the future. In general, the uncertainty appears to grow with larger prediction horizons. In the case of the red agent, the predictions slightly bifurcate around the obstacle. This clearly shows that the model is able to predict multiple potential futures of an agent (and the swarm) based on the inherent uncertainty of the task and environment. On the right, we see the probability distributions for discretized x- and y-coordinates of all agents. Again, some distributions are multimodal, which reinforces the insight that our predictions can produce multiple, diverse, and potentially conflicting future states.


\section{Conclusion}
In this paper, we presented a neural network architecture, called SwarmNet, which learns to predict and imitate behavior of an observed swarm. The network uses a combination of one-dimensional convolutions, graph convolutions, contextual inputs, along with a curriculum learning scheme to efficiently extract the swarm dynamics from positions and velocities of a set of agents. We showed that SwarmNet achieves high levels of prediction accuracy and that it can even be applied to nondeterministic and uncertain environments. For future work, we want to investigate how different application domains and tasks affect the sample complexity of the method, and the effectiveness of a decentralized implementation of the network.

\section{Acknowledgement}
This research was funded by a gift from the Intel Corporation. Partial funding was also provided by the Defense Advanced Research Projects Agency (DARPA) under Cooperative Agreement Number HR0011-18-2-0022.  The content of the information does not necessarily reflect the position or the policy of the Government, and no official endorsement should be inferred.  Approved for public release; distribution is unlimited.
\bibliographystyle{IEEEtran}
\bibliography{gnn}

\begin{thebibliography}{10}
\providecommand{\url}[1]{#1}
\csname url@rmstyle\endcsname
\providecommand{\newblock}{\relax}
\providecommand{\bibinfo}[2]{#2}
\providecommand\BIBentrySTDinterwordspacing{\spaceskip=0pt\relax}
\providecommand\BIBentryALTinterwordstretchfactor{4}
\providecommand\BIBentryALTinterwordspacing{\spaceskip=\fontdimen2\font plus
\BIBentryALTinterwordstretchfactor\fontdimen3\font minus
  \fontdimen4\font\relax}
\providecommand\BIBforeignlanguage[2]{{%
\expandafter\ifx\csname l@#1\endcsname\relax
\typeout{** WARNING: IEEEtran.bst: No hyphenation pattern has been}%
\typeout{** loaded for the language `#1'. Using the pattern for}%
\typeout{** the default language instead.}%
\else
\language=\csname l@#1\endcsname
\fi
#2}}

\bibitem{arai2002advances}
T.~Arai, E.~Pagello, L.~E. Parker, \emph{et~al.}, ``Advances in multi-robot
  systems,'' \emph{IEEE Transactions on robotics and automation}, vol.~18,
  no.~5, pp. 655--661, 2002.

\bibitem{lagoudakis2004simple}
M.~G. Lagoudakis, M.~Berhault, S.~Koenig, P.~Keskinocak, and A.~J. Kleywegt,
  ``Simple auctions with performance guarantees for multi-robot task
  allocation,'' in \emph{2004 IEEE/RSJ International Conference on Intelligent
  Robots and Systems (IROS)(IEEE Cat. No. 04CH37566)}, vol.~1.\hskip 1em plus
  0.5em minus 0.4em\relax IEEE, 2004, pp. 698--705.

\bibitem{Reynolds1999}
C.~W. Reynolds, ``{Steering behaviors for autonomous characters},'' \emph{Game
  Developers Conference}, pp. 763--782, 1999.

\bibitem{schaal1999imitation}
S.~Schaal, ``Is imitation learning the route to humanoid robots?'' \emph{Trends
  in cognitive sciences}, vol.~3, no.~6, pp. 233--242, 1999.

\bibitem{osa2018algorithmic}
T.~Osa, J.~Pajarinen, G.~Neumann, J.~A. Bagnell, P.~Abbeel, J.~Peters,
  \emph{et~al.}, ``An algorithmic perspective on imitation learning,''
  \emph{Foundations and Trends{\textregistered} in Robotics}, vol.~7, no. 1-2,
  pp. 1--179, 2018.

\bibitem{Helbing2000}
D.~Helbing, I.~Farkas, and T.~Vicsek, ``{Simulating dynamical features of
  escape panic},'' \emph{Nature}, vol. 407, no. 6803, pp. 487--490, 2000.

\bibitem{cortes2017coordinated}
J.~Cort{\'e}s and M.~Egerstedt, ``Coordinated control of multi-robot systems: A
  survey,'' \emph{SICE Journal of Control, Measurement, and System
  Integration}, vol.~10, no.~6, pp. 495--503, 2017.

\bibitem{mesbahi2010graph}
M.~Mesbahi and M.~Egerstedt, \emph{Graph theoretic methods in multiagent
  networks}.\hskip 1em plus 0.5em minus 0.4em\relax Princeton University Press,
  2010, vol.~33.

\bibitem{chernova2007multiagent}
S.~Chernova and M.~Veloso, ``Multiagent collaborative task learning through
  imitation,'' in \emph{Proceedings of the fourth International Symposium on
  Imitation in Animals and Artifacts}, 2007, pp. 74--79.

\bibitem{zhan2018generative}
E.~Zhan, S.~Zheng, Y.~Yue, and P.~Lucey, ``Generative multi-agent behavioral
  cloning,'' \emph{arXiv preprint arXiv:1803.07612}, 2018.

\bibitem{huttenrauch2017guided}
M.~H{\"u}ttenrauch, A.~{\v{S}}o{\v{s}}i{\'c}, and G.~Neumann, ``Guided deep
  reinforcement learning for swarm systems,'' \emph{arXiv preprint
  arXiv:1709.06011}, 2017.

\bibitem{sartoretti2018primal}
G.~Sartoretti, J.~Kerr, Y.~Shi, G.~Wagner, T.~Kumar, S.~Koenig, and H.~Choset,
  ``Primal: Pathfinding via reinforcement and imitation multi-agent learning,''
  \emph{arXiv preprint arXiv:1809.03531}, 2018.

\bibitem{Hocaouglu19weapon}
M.~F. Hocao{\u{g}}lu, ``Weapon target assignment optimization for land based
  multi-air defense systems: A goal programming approach,'' \emph{Computers \&
  Industrial Engineering}, vol. 128, pp. 681--689, 2019.

\bibitem{Kline18WTA}
A.~Kline, D.~Ahner, and R.~Hill, ``The weapon-target assignment problem,''
  \emph{Computers \& Operations Research}, 2018.

\bibitem{Guvenc18detection}
I.~Guvenc, F.~Koohifar, S.~Singh, M.~L. Sichitiu, and D.~Matolak, ``Detection,
  tracking, and interdiction for amateur drones,'' \emph{IEEE Communications
  Magazine}, vol.~56, no.~4, pp. 75--81, 2018.

\bibitem{hochreiter1997long}
S.~Hochreiter and J.~Schmidhuber, ``Long short-term memory,'' \emph{Neural
  computation}, vol.~9, no.~8, pp. 1735--1780, 1997.

\bibitem{Battaglia2018}
\BIBentryALTinterwordspacing
P.~W. Battaglia, J.~B. Hamrick, V.~Bapst, A.~Sanchez-Gonzalez, V.~Zambaldi,
  M.~Malinowski, A.~Tacchetti, D.~Raposo, A.~Santoro, R.~Faulkner, C.~Gulcehre,
  F.~Song, A.~Ballard, J.~Gilmer, G.~Dahl, A.~Vaswani, K.~Allen, C.~Nash,
  V.~Langston, C.~Dyer, N.~Heess, D.~Wierstra, P.~Kohli, M.~Botvinick,
  O.~Vinyals, Y.~Li, and R.~Pascanu, ``{Relational inductive biases, deep
  learning, and graph networks},'' 2018. [Online]. Available:
  \url{http://arxiv.org/abs/1806.01261}
\BIBentrySTDinterwordspacing

\bibitem{Gal2016Uncertainty}
Y.~Gal, ``Uncertainty in deep learning,'' Ph.D. dissertation, University of
  Cambridge, 2016.

\bibitem{Srivastava2014}
N.~Srivastava, G.~Hinton, A.~Krizhevsky, I.~Sutskever, and R.~Salakhutdinov,
  ``Dropout: A simple way to prevent neural networks from overfitting,''
  \emph{J. Mach. Learn. Res.}, vol.~15, no.~1, pp. 1929--1958, Jan. 2014.

\bibitem{Kipf2018}
T.~Kipf, E.~Fetaya, K.-c.~W. Max, and W.~Richard, ``{Neural Relational
  Inference for Interacting Systems},'' 2018.

\end{thebibliography}

\end{document}